%% file: main.tex
\documentclass[a4,journal]{IEEEtran}

\usepackage{amsmath,amsfonts}
\usepackage{algorithmic}
\usepackage{algorithm}
\usepackage{array}
\usepackage[caption=false,font=normalsize,labelfont=sf,textfont=sf]{subfig}
\usepackage{textcomp}
\usepackage{stfloats}
\usepackage{url}
\usepackage{verbatim}
\usepackage{graphicx}
\usepackage{cite}
\usepackage[table]{xcolor}

\usepackage{siunitx}
\usepackage{placeins}
\usepackage[Option]{overpic}
\usepackage{tabularx,booktabs}
\usepackage{multirow}
\usepackage{xcolor}
\usepackage[capitalise]{cleveref}
\usepackage[acronym]{glossaries}
\usepackage{tikz}
\usepackage{bm}

\newacronym{EEG}{EEG}{Electroencephalography}
\newacronym{EMG}{EMG}{Electromyography}
\newacronym{EOG}{EOG}{Electrooculography}
\newacronym{LLM}{LLM}{Large Language Model}
\newacronym{MCU}{MCU}{Microcontroller}
\newacronym{ML}{ML}{Machine Learning}
\newacronym{DVS}{DVS}{Dynamic Vision Sensor}
\newacronym{SiP}{SiP}{System-in-package}
\newacronym{SoC}{SoC}{System-on-Chip}
\newacronym{IC}{IC}{Integrated Circuit}
\newacronym{PSRAM}{PSRAM}{Pseudo-Static RAM}
\newacronym{PMIC}{PMIC}{Power Management Integrated Circuit}
\newacronym{FPC}{FPC}{Flexible Printed Circuit}
\newacronym{IMU}{IMU}{Inertial Measurement Unit}
\begin{document}

\title{OpenGlass: Ultra-Low-Power On-Device AI Eyewear with Event-based Vision}


\author{Pietro~Bonazzi,
        Julian~Moosmann,
        Ahmet~Celik,
        Philipp~Mayer,
        and~Michele~Magno,~\IEEEmembership{Fellow,~IEEE}%
\thanks{Pietro Bonazzi and Julian Moosmann contributed equally to this work.}%
\thanks{P.~Bonazzi, J.~Moosmann, A.~Celik, P.~Mayer, and M.~Magno are with the Department of Information Technology and Electrical Engineering, ETH~Z\"{u}rich, 8092~Z\"{u}rich, Switzerland (e-mail: pbonazzi@ethz.ch; mojulian@ethz.ch; celika@ethz.ch; mayerph@ethz.ch; mmagno@ethz.ch).}%
\thanks{Corresponding authors: P.~Bonazzi (pbonazzi@ethz.ch - Neural Network) and J.~Moosmann (jmoosmann@ethz.ch - System Architecture).}%
\thanks{Manuscript submitted June~5, 2026.}%
}

\markboth{IEEE INTERNET OF THINGS JOURNAL, Under Review, June~2026}%
{Bonazzi P. , Moosmann J. \MakeLowercase{\textit{et al.}}: A Sample Article Using IEEEtran.cls for IEEE Journals}


\maketitle

\input{00_abstract}


\input{01_introduction}
\input{02_relatedWork}
\input{03_HW_architecture}
\input{04_model}
\input{05_a_results_HW}
\input{05_b_results_model}
\input{05_c_results_system}
\input{06_discussion}
\input{07_conclusion}


\bibliography{IEEEabrv,./main}

\end{document}

%% file: 00_abstract.tex
\begin{abstract}
Smart eyewear enables unobtrusive, context-aware interaction through multimodal sensors and on-device intelligence, but is severely limited by power, memory, and compute constraints in a compact form factor. Open-hardware platforms supporting event-based vision and embedded ML at this scale are rare.
This work introduces an open-source smart glasses platform for rapid prototyping of novel sensors and algorithms. Its modular design uses a flexible FPC interposer to support both event-based and frame-based cameras without full PCB redesign. A hardware-software co-designed power management system combines a configurable PMIC with event-driven wake-up via an nRF5340 coordinator, keeping the GAP9 RISC-V SoC powered down between inferences. The prototype achieves up to 11.5 hours of continuous on-device ML from a 200 mAh battery.
As a demonstration, an egocentric hand gesture recognition pipeline was evaluated on the LynX dataset using polarity-separated event histograms from a Prophesee GENX320 camera. R(2+1)D achieved the best cross-subject accuracy of 83.94\% (macro F1 = 0.781) under leave-two-subjects-out validation, with 78.3 ms end-to-end inference latency on the GAP9. Temporal augmentation and removal of ambiguous classes provided the largest gains (+8.9 pp). All hardware designs, firmware, and models are released open source.
\end{abstract}

%% file: 01_introduction.tex
\section{Introduction}

\IEEEPARstart{S}{mart} eyewear has recently emerged as a compelling commercial wearable device, offering the potential for seamless, context-aware human–computer interaction. Positioned directly in the user’s line of sight, it enables not only natural visual augmentation but also the sensing and interpretation of what the user sees, hears, and—potentially—even smells or feels.~\cite{Hole2025_Social_Impact}

\begin{figure}[t]
    \centering
    \includegraphics[width=0.8\columnwidth]{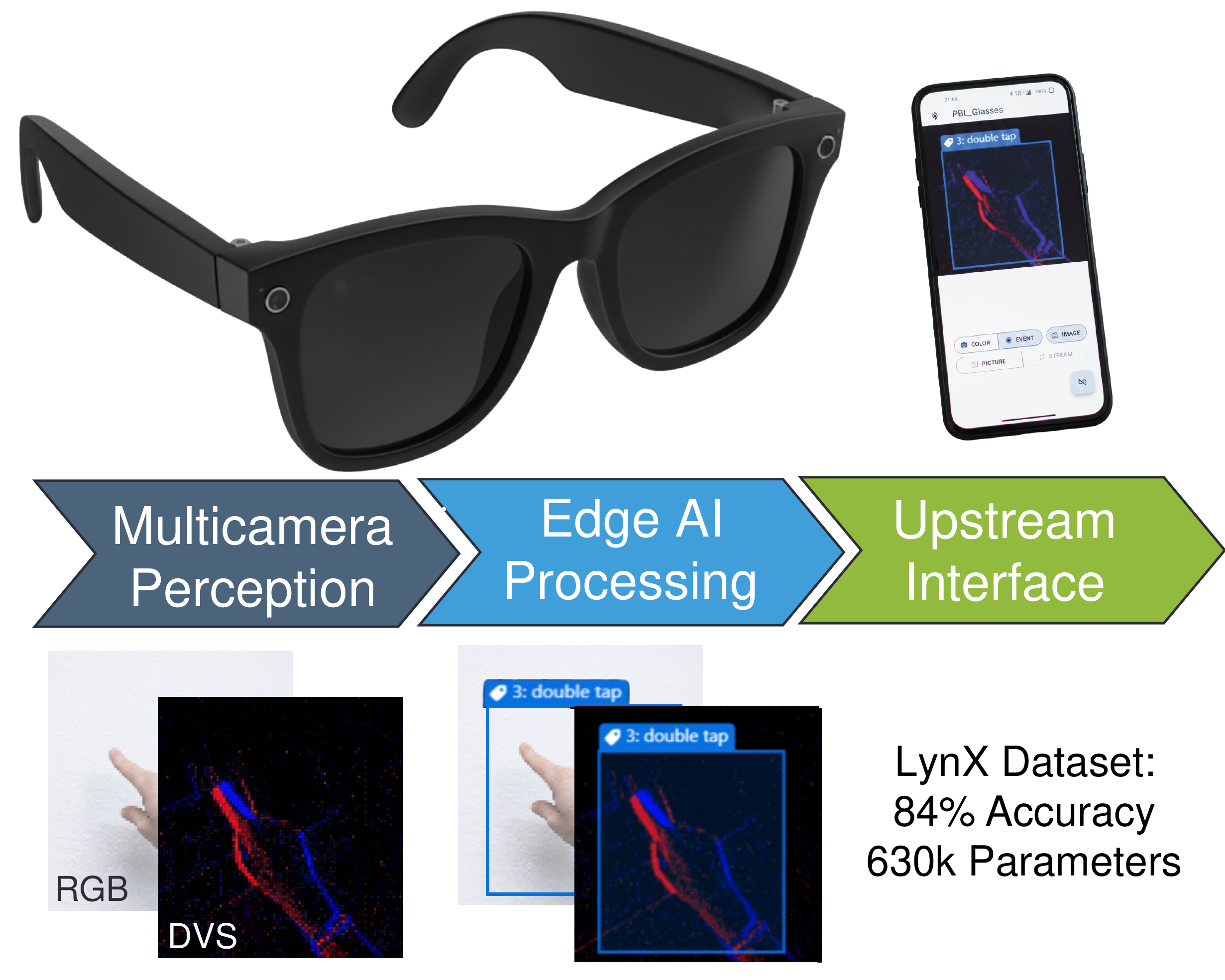}
    \caption{Overview of the OpenGlass platform. The system integrates multicamera perception (RGB + event-based DVS), on-device AI processing with the GAP9 SoC, and an upstream interface.}
    \label{fig:teaser}
\end{figure}

As a multimodal sensor hub, smart eyewear can generate highly personalized and context-rich data. Front-facing vision sensors perceive the environment and underpin augmented and mixed reality applications~\cite{goesele2025imagingalldaywearablesmart}. Eye-facing sensors track user attentiveness and cognitive load~\cite{Ghosh2024, Marinova2025}, while microphones enable not only voice control and sound recognition but also ambient awareness. Biopotential sensors such as \gls{EEG} provide deeper insight into a user’s mental and emotional states~\cite{Frey2025_GAPses, schärer2024electrasightsmartglassesfully}. Looking ahead, emerging modalities—including microwave radar and terahertz detectors—may further expand these capabilities.~\cite{Ma2025,Fu2022}

While such systems have the potential to revolutionize how humans perceive and interact with their environment, they also raise pressing concerns related to privacy, surveillance, and broader societal implications. It is therefore unsurprising that major advances in smart eyewear are currently driven by Big Tech companies—actors whose business models often rely on extensive data collection and targeted advertising.~\cite{Stephanidis2025_HCIchallanges, Hole2025_Social_Impact}

When combined with recent breakthroughs in \glspl{LLM}, the tight integration between user and device opens the door to a new generation of intelligent, always-on agents. These systems move beyond simple voice or text interfaces toward an assistive embodiment of AI—a digital co-pilot that continuously senses both the user and their environment, adapts in real time, and can even guide behaviour. Over time, this may evolve into a truly context-aware companion, seamlessly integrated into our cognitive and perceptual experience.~\cite{Waisberg2024, GOOGLE2025_VRandLLM, Wang2025}

Advancing toward this vision presents a complex roadmap of technical challenges. Smart eyewear must integrate diverse, high-bandwidth sensors and perform real-time, on-device inference using computationally and memory-intensive ML models—while remaining power-efficient, unobtrusive, and lightweight.~\cite{Wiese_2025} Ensuring responsiveness and preserving data privacy requires that much of this processing occur locally, which places substantial demands on embedded compute and memory. Meeting these constraints necessitates tight co-design across hardware, algorithms, and power management—through low-power processors, model compression, adaptive sampling, and event-driven computation~\cite{Flamand_2018_GAP8, Garofalo_2020_PULPNN, Rusci_2020_Quantization, Banbury_2021_MLPerfTiny, Lin_2020_MCUNet, David_2021_TFLiteMicro}. These trade-offs directly impact comfort, aesthetics, and the ability to support truly pervasive, always-on operation. Encouragingly, a new generation of microcontrollers , many built on the open and extensible RISC-V ISA, integrating parallel compute clusters and dedicated AI accelerator cores, is beginning to close this gap~\cite{Flamand_2018_GAP8, Garofalo_2020_PULPNN}. The openness of RISC-V further enables custom instruction set extensions tailored to inference workloads, making these platforms particularly promising for the tight co-design that wearable edge AI demands.

Among the sensing modalities available in smart eyewear, the event-based camera~\cite{Gallego_2022} represents a largely unexplored yet uniquely promising modality for always-on wearable sensing. Unlike frame-based cameras capturing redundant full-scene images at fixed intervals, event cameras respond asynchronously to luminance changes, emitting sparse, microsecond-resolution data only when the scene changes~\cite{EventCameraSurvey_2024}. This yields dramatic power savings during the static conditions that dominate daily wear, while preserving high temporal resolution when motion occurs. The native sparsity of the event stream also aligns naturally with efficient processing pipelines, enabling privacy-preserving on-device inference ~\cite{Lin_2020_MCUNet} . For augmented reality, attention tracking, and ambient scene understanding, the event camera thus offers a compelling combination of energy efficiency, temporal acuity, and robustness to high-dynamic-range outdoor lighting.

This work introduces an open-hardware platform designed to support rapid prototyping, evaluation, and deployment of novel sensors for smart eyewear systems. Fully open source and freely adaptable, the platform enables experimentation at the intersection of embedded intelligence, hardware design, and human–machine interaction. As a reference implementation, a compact, ultra-low-power smart glasses prototype is presented featuring both event-based~\cite{Lichtsteiner_2008_DVS, EventCameraSurvey_2024} and frame-based imaging sensors. Its utility is demonstrated through an on-device hand gesture recognition pipeline evaluated on the LynX dataset~\cite{LynX}, an egocentric event-based gesture collection captured in extended reality scenarios.

The key contributions of this work are:
\begin{enumerate}
    \item An open-source, modular smart glasses prototype in the Ray-Ban Stories form factor, integrating a \textbf{Prophesee GENX320 event-based camera} alongside a frame-based imager in a fully self-contained wearable—the first open platform to adopt event-based vision as the primary sensing modality. A flexible \gls{FPC} interposer decouples sensor selection from the main board, enabling plug-and-play addition of custom imagers and embedded ML algorithms without hardware redesign.

    \item A \textbf{novel event-based perception pipeline} for always-on wearable inference: asynchronous event streams are accumulated into polarity-separated temporal histograms and processed by a spatiotemporal R(2+1)D convolutional network adapted for event-histogram inputs. A stochastic temporal and polarity augmentation strategy is introduced, yielding an 8.9 percentage-point accuracy gain and demonstrating that standard video architectures can be effectively repurposed for event-based egocentric gesture recognition on microcontroller-class hardware.

    \item A hardware--software co-designed power management strategy, achieving up to 11.5\,hours of continuous on-device ML from an ultra-compact \qty{200}{\milli\ampere\hour} battery, demonstrating viability for full-day wearable deployment.

    \item A comprehensive system validation on the LynX egocentric event-based gesture dataset under a strict leave-two-subjects-out cross-subject protocol, achieving 83.94\% accuracy (macro F1\,=\,0.781) with 78.3\,ms inference latency on the GAP9, confirming end-to-end suitability for real-world embedded vision.
\end{enumerate}

The remainder of this paper is organized as follows: Section~\ref{section:relatedWork} reviews related commercial and research platforms in smart eyewear. Section~\ref{section:architecture} presents the proposed system architecture and hardware design. Section~\ref{section:architecture} details the data preprocessing pipeline and benchmark model architectures. Section~\ref{section:results_model} reports the network evaluation results and end-to-end system performance. Finally, Section~\ref{section:discussion} discusses the findings and limitations, and Section~\ref{section:conclusion} concludes the paper.

%% file: 02_relatedWork.tex
\begin{table*}[b]
\caption{Smart glasses platforms: a comparative overview of commercial products and open-hardware initiatives.}
\centering
\renewcommand{\arraystretch}{1.35}

\resizebox{\textwidth}{!}{%
\begin{tabular}{@{}l*{5}{c}@{}}
\toprule
 & Ray-Ban Meta Gen2 \cite{Meta_RayBan_Meta_Wayfarer_Gen2} & HoloLens 2 \cite{Microsoft_HoloLens2_Hardware} & Brilliant Labs – Frame \cite{Brilliant_Frame_Hardware} & Mentra - OSSG \cite{teamopensg} & \bfseries This Work \\
\midrule
Type & Industry & Industry & Industirial/Community & Community & Research \\
Weight & \qtyrange{51}{53}{\gram} & \qty{566}{\gram} & \qty{39}{\gram} & \qty{45}{\gram} & \qty{40}{\gram} \\

\multirow{2}{*}{Processor} & Snapdragon & Snapdragon 850 & LIFCL-17 & Cortex-M4 & GAP9 (RISC-V) \\[-1.0ex]
& AR1 Gen1 & & Cortex-M4 &  & Cortex-M33 \\

\multirow{2}{*}{Memory} & \qty{2}{\gibi\byte} RAM & \qty{4}{\gibi\byte} RAM & \qty{0.256}{\kibi\byte} + \qty{2.56}{\mebi\byte} RAM & \qty{256}{\kibi\byte} RAM & \qty{64}{\mebi\byte} RAM \\[-1.0ex]
& \qty{32}{\gibi\byte} Flash & \qty{64}{\gibi\byte} Flash & \qty{1}{\mebi\byte} Flash &  & \qty{256}{\mebi\byte} Flash \\

Runtime & \qty{8}{\hour} & \qty{3}{\hour} & \qtyrange{5}{6}{\hour} & \qty{15}{\hour} & \qty{11.5}{\hour} \\

\multirow{2}{*}{Camera type} & RGB \qty{12}{MP} & RGB \qty{8}{MP} & RGB \qty{0.92}{MP} & — & RGB \qty{0.3}{MP} \\

&   & Depth \qty{1}{MP} &  &  & DVS \qty{0.1}{MP}  \\

Other sensors & IMU, mic, touchpad & head \& eye tracking, IMU & IMU, mic, display & IMU, mic & IMU, mic \\

\textbf{Connectivity} & Bluetooth, Wi-Fi & Wi-Fi, Bluetooth, USB-C & BLE & BLE & BLE \\

System openness & Closed & Closed & Open-source & Open-source & Open-source \\

Availability & Commercial & Commercial & GitHub & GitHub & GitHub\scriptsize{$^a$} \\
\bottomrule
\end{tabular}
}
\scriptsize{\\ \vspace{3px} \hspace{380px} $^a$ Will be made available upon publication.}\\
\label{table:smartglasses_comparison}
\end{table*}

\section{Related Work}
\label{section:relatedWork}
With the growing market potential of wearable computing, the number of commercial smart eyewear products is rapidly increasing \cite{Wang2025}. Current developments can be broadly categorized into cloud-assisted commercial systems and research prototypes for sensor evaluation. An overview of representative commercial solutions and open-hardware initiatives is provided in \cref{table:smartglasses_comparison}.

Commercial products such as the \textsc{Ray-Ban Meta Gen2} rely heavily on smartphone and cloud connectivity for data processing. Integrated high-resolution imagers further increase computational and communication bandwidth demands. While these architectures enable advanced functionalities, they also raise privacy concerns and limit battery life to approximately \qty{8}{\hour} in the latest models.

In contrast, academic efforts have focused on alternative sensing modalities (e.g., \gls*{EEG} and eye tracking) and emphasize on-device inference to enhance privacy, reduce latency, and improve energy efficiency. Frey \textit{et al.} \cite{Frey2025_GAPses} demonstrated the integration of \gls{EEG} and \gls{EOG} sensing using dry electrodes in smart eyewear, achieving on-device subject and eye-movement classification with an operational lifetime exceeding \qty{22}{\hour} from a \qty{75}{\milli\ampere\hour} battery.

Schärer \textit{et al.} \cite{schärer2024electrasightsmartglassesfully} proposed a hybrid \gls{EOG} approach that combines contact and contactless electrodes in a smart-glasses form factor, enabling on-device classification with a latency of approximately \qty{60}{\milli\second} and an operational lifetime of up to \qty{3}{\day} from a \qty{175}{\milli\ampere\hour} battery.

\cite{moosmann2023ultraefficientondeviceobjectdetection} demonstrated the integration of a RISC-V processor into a \textsc{Ray-Ban Stories} frame, employing a single low-power RGB camera for on-device, real-time object detection. By executing a memory-optimized sub-million-parameter YOLO model, the system achieved an end-to-end latency of \qty{56}{\milli\second} (\qty{18}{fps}) from image capture to prediction, with a total power consumption of \qty{62.9}{\milli\watt}. This corresponds to approximately \qty{9.3}{\hour} of continuous operation with a \qty{154}{\milli\ampere\hour} battery.

Open hardware platforms that support the integration of new sensors remain rare, with only a few examples listed in \cref{table:smartglasses_comparison}. 
\textsc{Brilliant Labs} developed the \textsc{Frame} \cite{Brilliant_Frame_Hardware}, an open hardware/software concept providing public access to schematics, mechanical designs, and firmware. 
Its electronics combine an FPGA for image acquisition and display driving with a \textsc{Cortex-M4}–based \gls{SoC} for wireless communication and system management. 

Another example is the \textsc{OSSG} platform \cite{teamopensg}, developed by the \textsc{Team Open Smart Glasses} community. It offers an open-source hardware and software framework designed for experimentation with wearable sensing in an eyewear form factor. The electronics integrate a low-power \textsc{Cortex-M4} microcontroller with BLE connectivity and an \gls{IMU}, targeting developers and researchers interested in extending the smart-glasses ecosystem with new sensors and functionalities.

Beyond smart glasses, open wearable research platforms have demonstrated that compact, energy-harvesting or battery-powered hardware nodes can support on-device AI inference at sub-\qty{10}{\milli\watt} power levels~\cite{Polonelli_2021_HWatch}.

Event cameras have attracted increasing interest as sensing modalities for wearable, always-on systems~\cite{Lichtsteiner_2008_DVS, EventCameraSurvey_2024, Adra_2025_EventHumanReview}. Their asynchronous, sparse output reduces bandwidth and power compared to conventional frame-based sensors, making them particularly suitable for egocentric interaction~\cite{Damen_2018_EPICKitchens, Plizzari_2022_E2GOMOTION, Wang_2025}. Gesture recognition is a key application, with recent systems targeting always-on inference on event-sensor-based wearables~\cite{Bhattacharyya_2024, bhattacharyya2025helios20robustultralow}. To support egocentric event-based gesture research, the LynX dataset~\cite{LynX} provides annotated first-person event streams for extended reality interaction, and is used as the evaluation benchmark in this work. Concurrent work has also explored running lightweight egocentric action recognition models in real time directly on resource-constrained smart eyewear hardware~\cite{Santambrogio_2025_EgoSmartEyewear}.

To the best of our knowledge, this is the first fully open-source smart glasses platform to adopt an event-based camera as the primary sensing modality and to co-design both the sensing pipeline and the inference algorithm for that modality. Prior open platforms either fix the sensor type or provide only a compute substrate without an accompanying algorithmic pipeline. This work contributes hardware, a complete event-to-prediction software stack—covering asynchronous event accumulation, polarity-separated histogram preprocessing, spatiotemporal deep learning, and BLE result transmission—and an end-to-end evaluation under cross-subject conditions, collectively lowering the barrier for research at the intersection of wearable event-based perception and embedded AI.

%% file: 03_HW_architecture.tex

\begin{figure*}[ht]
    \centering
    \includegraphics[width=1\textwidth]{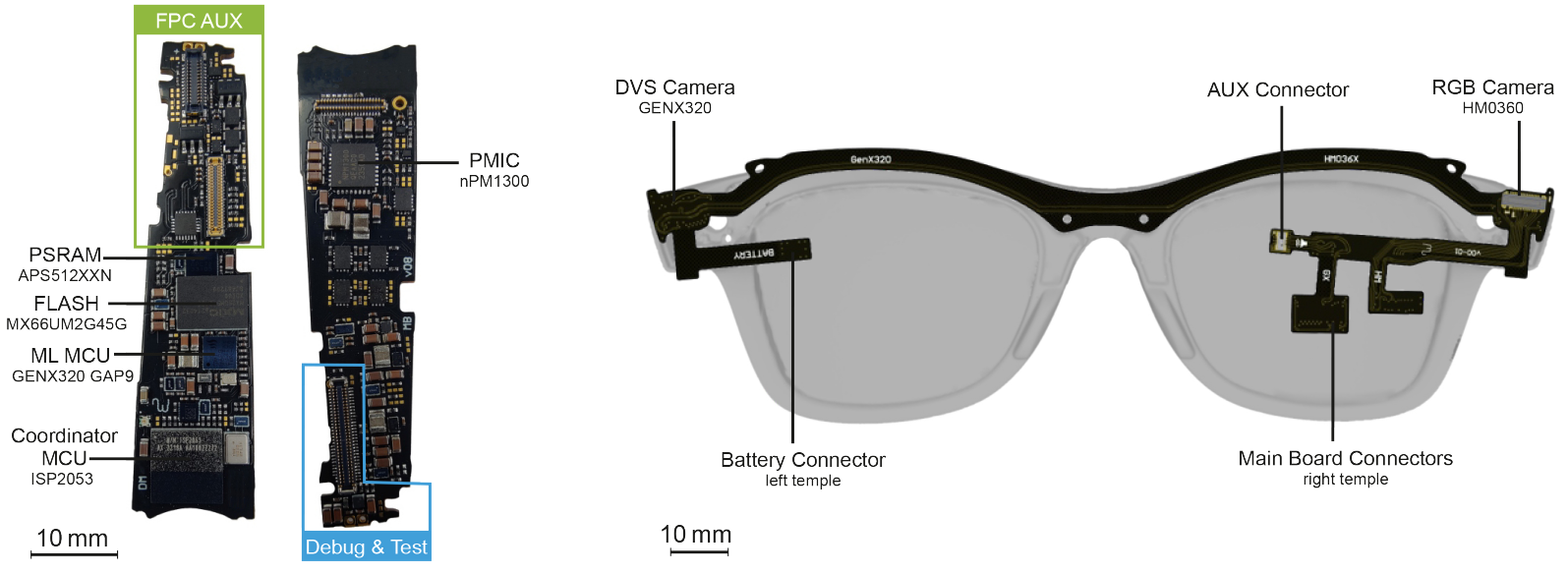}
    \caption{  Annotated photographs of the fabricated hardware. Left: the Main Board PCB (portrait view) with the FPC AUX attached,  showing the GAP9 ML MCU, nRF5340 coordinator, PSRAM, Flash, and nPM1300 PMIC. Right: the assembled smart glasses frame with the Prophesee GENX320 DVS camera (left front), HIMAX HM0360 RGB camera (right front), battery connector (left temple), and  Main Board connectors (right temple).}
    \label{fig:system_overview}
\end{figure*}

\section{System Architecture}
\label{section:architecture}
The hardware architecture of the proposed smart eyewear platform has been developed to achieve a high degree of integration while ensuring manufacturability, testability, and extensibility. The system is organized into three main components:
\\\textit{Main Board} — Housed within the right temple\footnote{from a user's ego-perspective} of the eyewear, this PCB integrates both a system coordinator and a machine learning microcontroller. This combination enables efficient on-device processing, wireless communication, and energy-efficient operation across dynamic application scenarios.
\\\textit{Frame FPC} — A flexible sensor carrier that connects the Main Board to two front-facing cameras and a lithium-ion polymer battery pack (\qty{200}{mAh}, $\qty{774}{mWh} \approx \qty{2.786}{kJ}$) located in the left temple. Its design supports a compact and lightweight form factor suitable for integration into eyewear frames, while keeping the system flexible with respect to sensor selection and placement.
\\\textit{Development Panel} — Designed to support testing and debugging, this panel provides standard programming interfaces and a USB port for wired data readout\Cref{fig:development} .\Cref{fig:system_overview} illustrates the overall system architecture, which is described in more detail in the following paragraphs.

\begin{figure}[h]
    \centering
    \includegraphics[width=1 \linewidth]{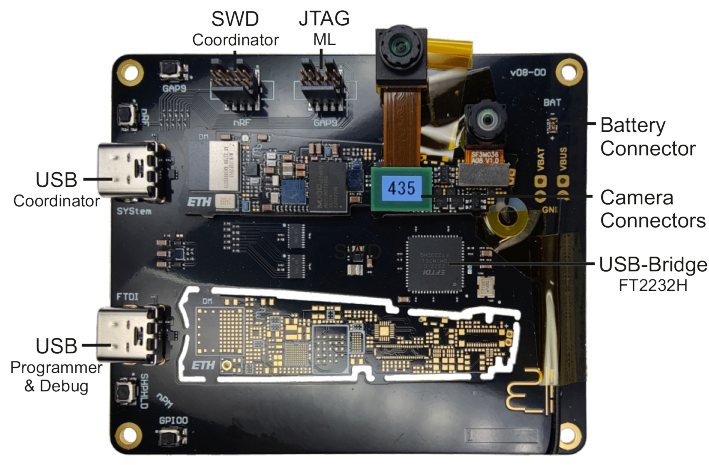}
    \caption{Development Panel used for prototyping and characterisation, providing USB connectivity for both the coordinator (nRF5340) and ML MCU (GAP9) via dedicated SWD and JTAG interfaces, an FT2232H USB-Bridge for wired data readout, and break-out connectors for battery and camera modules}
    \label{fig:development board}
\end{figure}

\subsection{Circuit overview}
\begin{figure}[ht]
    \centering
    \includegraphics[
        trim={0 0 10cm 0},
        clip,
        width=\linewidth       
    ]{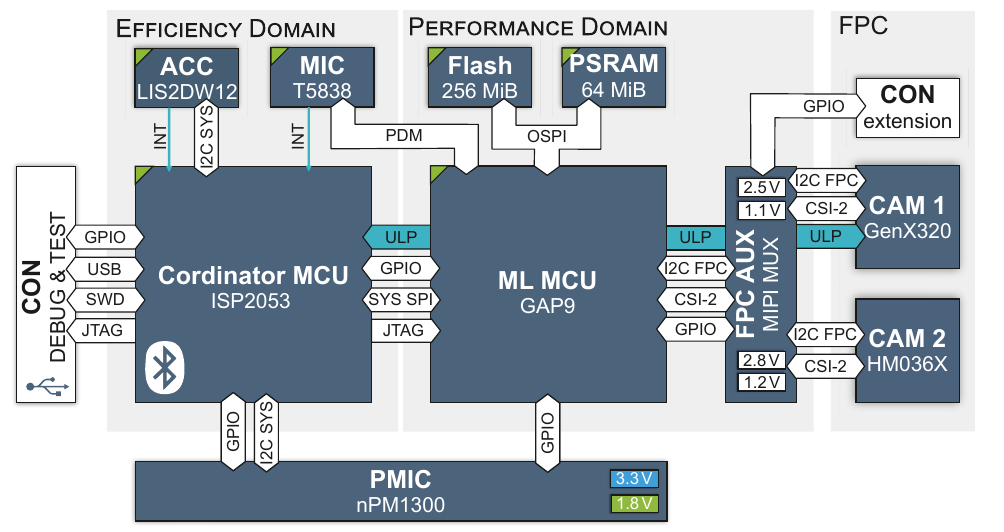}
    \caption{Block diagram of the proposed smart eyewear platform. The efficiency domain (nRF5340 coordinator, LIS2DW12 accelerometer, T5838 microphone) handles wireless communication and wake-up sensing; the performance domain (GAP9 ML MCU, 256\,MiB Flash,64\,MiB PSRAM) runs the vision pipeline. Camera modules (Prophesee GENX320 DVS and HIMAX HM0360 RGB) and auxiliary circuitry are hosted on the FPC board; power management is centralised in the nPM1300 PMIC.}
    \label{fig:block_diagram}
\end{figure}

The circuit consists of two main functional domains: an efficiency domain that enables low-leakage operation, and a performance domain that supports high-bandwidth data acquisition and on-board inference, as shown in \cref{fig:block_diagram}. This strict separation, combined with fine-grained power gating of individual functional blocks, minimizes application-specific leakage currents and fosters duty-cycled or event-driven operation. Battery charging, system monitoring, and power supply management are handled by a software-configurable \gls{PMIC} (nPM1300).

\subsubsection{Efficiency Domain} The complete system is coordinated by the \gls{SiP} ISP2053 from \textsc{InsightSiP} built around the \textsc{Nordic Semiconductor} nRF5340 microcontroller, which integrates a \qty{32}{bit} \textsc{Cortex}-M33 host core operating at \qty{128}{\mega\hertz} with Bluetooth~5.4 support. It is used for system configuration, power management, and orchestrates the wireless data transmission. Activatable through TPS22922 load switches ($I_{Q_{OFF}} = \qty{50}{\nano\ampere}$ at \SI{1.8}{\volt}), the efficiency domain directly incorporates two sensors: the LIS2DW12 accelerometer and the T5838 microphone. They can be configured as event-driven wake-up sources for motion and acoustic activity, with a quiescent current of $I_{Q} = \qty{380}{\nano\ampere}$ and $I_{Q} = \qty{20}{\micro\ampere}$ at \SI{1.8}{\volt}, respectively.

\subsection{Performance Domain} The performance domain is centered on the \textsc{GreenWaves Technology} GAP9 \gls{SoC}~\cite{Flamand_2018_GAP8, Muller_2024_GAP9Shield}, which integrates a nine-core RISC-V compute cluster equipped with an AI accelerator (NE16)~\cite{Tortorella_2022_RedMulE, Garofalo_2020_PULPNN} and transprecision floating-point support (IEEE 32-bit, 16-bit, and bfloat16)~\cite{Prasad_2024_Siracusa}. This architecture achieves up to \qty{330}{\micro\watt\per GOP} energy efficiency, delivering \qty{15.6}{GOPs} for DSP workloads and \qty{32.2}{GMACs} for machine learning tasks. A dedicated single-core RISC-V controller manages peripherals and delegates workloads to the cluster. The \gls{IC} also provides high-bandwidth interfaces for audio and video. Energy efficiency is further improved through dynamic frequency scaling (up to \qty{370}{\mega\hertz}), voltage scaling, and automatic clock gating, enabling application-specific optimization. The \gls{SoC} integrates \qty{1.6}{\mega\byte} of L2 RAM and \qty{2}{\mega\byte} of in-package non-volatile memory, complemented by external \SI{64}{\mebi\byte} of \gls{PSRAM} and \SI{256}{\mebi\byte} of NOR flash.

In addition to the ML MCU, the performance domain includes the FPC auxiliary circuits responsible for power-gating, supplying, and clocking the camera modules and accessories mounted on the FPC board.
The \textsc{Prophesee} GENX320 event camera is powered by \qty{2.5}{\volt} (P5900TLX-2.5), \qty{1.8}{\volt} (TPS22922), and \qty{1.1}{\volt} (TPS62841). Start-up sequencing and ultra-low-power configuration are managed by an LM3880 power-sequencing IC. Similarly, the \textsc{HIMAX} HM0360 RGB camera is powered by \qty{2.8}{\volt} (P5900TLX-2.8), \qty{1.8}{\volt} (TPS22922), and \qty{1.2}{\volt} (TPS62841).


%% file: 04_model.tex
\section{Neural Network}\label{sec:model}
The data pre-processing pipeline strictly follows the procedure outline in the Lynx Dataset~\cite{LynX}. Furthermore, a collection of convolution-based architectures compatible with the GAP9 SoC is explored to perform the hand-gesture recognition task used to demostrate the platform capabilities.

\subsection{Data Pre-Processing}
The GENX320 outputs asynchronous $  (x, y, t, p)  $ events~\cite{Gallego_2022, Lichtsteiner_2008_DVS}.
Events are accumulated into polarity-separated histogram frames of size $64{\times}64$ over fixed time windows~\cite{Sironi_2018_HATS}.
Each frame thus has two channels: accumulated positive and negative event counts, normalized to $  [0,1]  $.
Gesture clips are formed by stacking $  T{=}10  $ consecutive frames with a temporal stride of~5 frames between the start of consecutive clips.
This yields input tensors of shape $2{\times}10{\times}64{\times}64$ (channels~$  \times  $~time~$  \times  $~height~$  \times  $~width), matching the expected input format of all evaluated architectures.
To improve generalization across subjects stochastic augmentations are applied at training time:
\begin{itemize}
\item \textit{Temporal shift}: the clip is shifted by up to $  \pm8  $ frames along the time axis.
\item \textit{Temporal scaling}: the clip duration is rescaled by a random factor drawn from $  [0.85,\,1.15]  $.
\item \textit{Polarity dropout}: all events of a randomly chosen polarity channel are zeroed out with probability 0.2.
\item \textit{Frame dropout}: individual frames within the clip are dropped with probability 0.3.
\item \textit{Horizontal flip}: the spatial frame is mirrored left-to-right with probability 0.5.
\end{itemize}

\subsection{Benchmark Model Architectures}
The gesture recognition pipeline processes event streams from the \textsc{Prophesee} GENX320 sensor~\cite{prophesee_genx320}.
Raw events are accumulated into two-channel polarity frames (positive and negative events separately) at a spatial resolution of $64{\times}64$ pixels, yielding compact representations compatible with the memory constraints of the GAP9 \gls{SoC}.
Temporal clips of $  T{=}10  $ consecutive frames are extracted with a stride of~5, providing a 50\% temporal overlap between successive clips.
Five backbone architectures are evaluated, all constrained to sub-million parameters to fit within the \qty{1.6}{\mega\byte} L2 RAM of the GAP9 cluster (MobileNet is the single exception, included for reference)~\cite{Banbury_2021_MLPerfTiny, David_2021_TFLiteMicro, Lin_2020_MCUNet}:
\paragraph{R(2+1)D}~\cite{Tran_2018_r21d} factorizes 3D spatiotemporal convolutions into a 2D spatial convolution followed by a 1D temporal convolution, building on foundational 3D CNN architectures~\cite{Tran_2015_C3D, Carreira_2017_I3D}.
This decomposition reduces the parameter count relative to full 3D convolutions while preserving both spatial appearance and motion information.
With 627,407 parameters, it achieves the best overall accuracy in this study.
\paragraph{TSM} (Temporal Shift Module)~\cite{Lin_2019_tsm} inserts a zero-cost channel-shift operation into a ResNet backbone, enabling temporal information exchange across frames without additional parameters.
The shift is applied before the standard 2D residual blocks, keeping the architecture deployable on hardware without native 3D convolution support (589,887 parameters).
\paragraph{TCN} (Temporal Convolutional Network)~\cite{Bai_2018_tcn} models the sequence of spatial features extracted per frame using stacked dilated causal convolutions, achieving a large receptive field over the temporal axis with fewer parameters (493,137) than recurrent alternatives. TCNs have demonstrated strong accuracy-efficiency trade-offs on resource-constrained wearable hardware~\cite{Ingolfsson_2021_ECGTCN}.
\paragraph{DD-TCN} (Dilated Depthwise TCN) extends the TCN design with depthwise separable dilated convolutions, trading a modest accuracy drop for improved multiply-accumulate efficiency (526,545 parameters).
\paragraph{MobileNet}~\cite{Howard_2017_mobilenet} applies depthwise separable 2D/3D convolutions.
Although the architecture is designed for mobile deployment, the adopted configuration carries 3,776,855 parameters—the heaviest model evaluated—and does not outperform the leaner R(2+1)D backbone.

\subsection{Training Procedure}
All models are trained end-to-end with a multi-task objective combining a cross-entropy classification loss $  \mathcal{L}_\text{cls}  $ and a bounding-box regression loss $  \mathcal{L}_\text{bbox}  $.
The Adam optimizer is used with an initial learning rate $  \eta{=}10^{-3}  $ and weight decay $  \lambda{=}10^{-4}  $, a cosine-annealing schedule, and a batch size of 64.
Training runs for up to 100 epochs with early stopping (patience~15).
\paragraph{Class imbalance.}
The LynX dataset~\cite{LynX} exhibits significant label imbalance across gesture classes.
Two complementary mitigation strategies are evaluated:
\begin{enumerate}
\item \textit{Undersampling}: each class is capped at a fixed number of samples, equalizing class frequencies before training.
\item \textit{Drop g12/g13}: gesture classes 12 and 13, which are the most ambiguous and underrepresented, are removed entirely, reducing the classification head to 11 output neurons.
\end{enumerate}
\paragraph{Bundled training.}
For the final reported model the training and validation splits are merged into a single training set after hyperparameter selection, and the model is retrained from scratch using the combined data before evaluation on the held-out test subjects.

%% file: 05_a_results_HW.tex


%% file: 05_b_results_model.tex
\section{Network Evaluation}
\label{section:results_model}

\subsection{Evaluation Protocol}

Models are evaluated under a strict leave-two-subjects-out protocol: subjects 8 and 9 are held out as the test set ($N{=}1{,}040$ samples), and no data from these subjects appears at any stage of training or hyperparameter selection. This design ensures that reported scores reflect genuine cross-subject generalisation rather than in-distribution performance.

Three complementary metrics are reported. \textit{Test accuracy} measures the proportion of correctly classified gestures and provides an overall performance figure. \textit{Macro-averaged F1} weights each class equally, making it sensitive to performance on underrepresented classes regardless of their support. \textit{Weighted F1} weights each class by its sample count, tracking overall discriminative quality. Together, the gap between macro and weighted F1 quantifies how much class imbalance degrades per-class performance.

\subsection{Architecture Comparison}

Table~\ref{tab:arch_comparison} compares the five evaluated architectures at clip length $T{=}10$ and stride~5. Three groups of performance are visible. MobileNet (61.79\%) achieves the lowest accuracy despite carrying nearly $6\times$ more parameters than the next-largest model, indicating that parameter count alone does not compensate for an insufficient temporal inductive bias when processing multi-frame event histogram clips. The dilated convolutional models—DD-TCN (63.79\%) and TCN (69.06\%)—improve substantially by modelling temporal dependencies across the clip, with the standard TCN outperforming the depthwise-separable variant, suggesting that the separability approximation introduces a capacity bottleneck at this scale. TSM (79.62\%) benefits from temporal information exchange across frames without additional parameters, confirming that even a lightweight temporal mechanism yields large gains over frame-independent processing. R(2+1)D (83.94\%) attains the highest accuracy, outperforming TSM by 4.3 percentage points, and achieves this with fewer parameters than MobileNet.

\begin{table}[ht]
\centering
\caption{Best test results per architecture (strict subject hold-out, $T{=}10$, stride~5).}
\label{tab:arch_comparison}
\begin{tabular}{lcccc}
\toprule
\textbf{Architecture} & \textbf{\#Params} & \textbf{Test Acc.} & \textbf{Macro F1} & \textbf{Wtd.\ F1} \\
\midrule
MobileNet     & 3\,776\,855 & 61.79\%          & 0.486          & 0.609 \\
DD-TCN        & 526\,545    & 63.79\%          & 0.486          & 0.638 \\
TCN           & 493\,137    & 69.06\%          & 0.551          & 0.690 \\
TSM           & 589\,887    & 79.62\%          & 0.731          & 0.795 \\
\rowcolor{green!20}
R(2+1)D       & 627\,407    & \textbf{83.94\%} & \textbf{0.781} & \textbf{0.840} \\
\bottomrule
\end{tabular}
\end{table}

The consistent ranking across accuracy and both F1 variants suggests that class imbalance does not disproportionately favour any single architecture. The macro–weighted F1 gap is largest for the weakest model (0.123 for MobileNet) and narrows for stronger ones (0.059 for R(2+1)D), indicating that better architectures reduce, but do not eliminate, the performance penalty on minority classes.



\subsection{Augmentation and Training Strategy}

The winning R(2+1)D configuration excludes gesture classes 12 and 13 (the two most ambiguous and lowest-support classes in LynX), applies stochastic temporal and polarity augmentation, and is retrained from scratch on the combined train–validation split before final evaluation. Table~\ref{tab:aug_ablation} isolates the contribution of each strategy.

\begin{table}[!t]
\centering
\caption{Augmentation ablation for R(2+1)D (strict subject hold-out, $T{=}10$, stride~5). Each row adds one strategy cumulatively.}
\label{tab:aug_ablation}
\begin{tabular}{lccc}
\toprule
\textbf{Configuration} & \textbf{Test Acc.} & \textbf{Macro F1} & \textbf{Wtd.\ F1} \\
\midrule
Baseline (no aug, all classes)   & 71.72\% & 0.620 & 0.721 \\
$+$ drop g12/13 $+$ augmentation & 80.58\% & 0.770 & 0.814 \\
\rowcolor{green!20}
$+$ bundled training             & \textbf{83.94\%} & \textbf{0.781} & \textbf{0.840} \\
\bottomrule
\end{tabular}
\end{table}

Removing the two ambiguous classes and applying augmentation delivers the largest single gain (+8.86~pp accuracy, +0.150 macro F1). The improvement in macro F1 is proportionally larger than in accuracy, confirming that these strategies primarily benefit the tail of the class distribution. The additional +3.36~pp from bundled training reflects the value of every training sample in a small, 10-subject dataset: withholding a validation split for early stopping deprives the model of data that, once the optimal configuration is fixed, would improve generalisation.

To quantify robustness beyond the single held-out split, 6-fold cross-validation is performed with the best R(2+1)D configuration. Table~\ref{tab:cv_folds} reports per-fold results for both the drop-g12/13 and the undersample strategy. The drop-g12/13 variant yields a mean cross-validated accuracy of 74.55\%\,$\pm$\,6.07~pp, and the undersample variant yields 75.39\%\,$\pm$\,3.85~pp—both approximately 9~pp below the bundled hold-out result. The undersample strategy shows lower variance across folds, suggesting that equalising class frequencies improves stability at the cost of slightly lower macro F1 (0.654 vs. 0.702).

\begin{table}[!t]
\centering
\caption{Per-fold cross-validation results for the R(2+1)D backbone ($T{=}10$, stride~5, aug). Test subjects differ per fold; reported accuracy and macro F1 are on the held-out test subjects.}
\label{tab:cv_folds}
\setlength{\tabcolsep}{4pt}
\begin{tabular}{clcccc}
\toprule
\textbf{Strategy} & \textbf{Test subjects} & \textbf{Acc.} & \textbf{Macro F1} & \textbf{Best ep.} \\
\midrule
\multirow{6}{*}{\rotatebox[origin=c]{90}{drop g12/13}}
 & 8, 9   & 81.73\% & 0.766 & 18 \\
 & 5, 10  & 75.00\% & 0.704 & 38 \\
 & 12, 16 & 63.28\% & 0.595 & 49 \\
 & 6, 15  & 76.08\% & 0.743 & 25 \\
 & 13, 14 & 74.83\% & 0.681 & 31 \\
 & 4, 7   & 76.36\% & 0.726 & 14 \\
\cmidrule{2-5}
 & \textbf{mean $\pm$ std} & \textbf{74.55 $\pm$ 6.07} & \textbf{0.702 $\pm$ 0.060} & \\
\midrule
\multirow{6}{*}{\rotatebox[origin=c]{90}{undersample}}
 & 8, 9   & 82.61\% & 0.724 & 15 \\
 & 5, 10  & 72.72\% & 0.609 & 17 \\
 & 12, 16 & 72.24\% & 0.576 & 31 \\
 & 6, 15  & 75.13\% & 0.687 & 38 \\
 & 13, 14 & 76.32\% & 0.641 & 59 \\
 & 4, 7   & 73.39\% & 0.685 & 47 \\
\cmidrule{2-5}
 & \textbf{mean $\pm$ std} & \textbf{75.40 $\pm$ 3.85} & \textbf{0.654 $\pm$ 0.055} & \\
\bottomrule
\end{tabular}
\end{table}

A striking feature of the fold-level results is that subjects 12 and 16 (fold 3) consistently represent the hardest test pair, achieving 63.28\% (drop g12/13) and 72.24\% (undersample)—the lowest in both configurations. This suggests that these two subjects exhibit gesture styles that diverge from the rest of the cohort, and that the overall cross-subject difficulty is driven by a small number of outlier subjects rather than uniform between-subject variability. The cross-validated figures are therefore a more conservative generalisation estimate than the primary hold-out result.

\subsection{Per-Class Analysis}

\begin{figure}[ht]
    \centering
    \includegraphics[width=\linewidth]{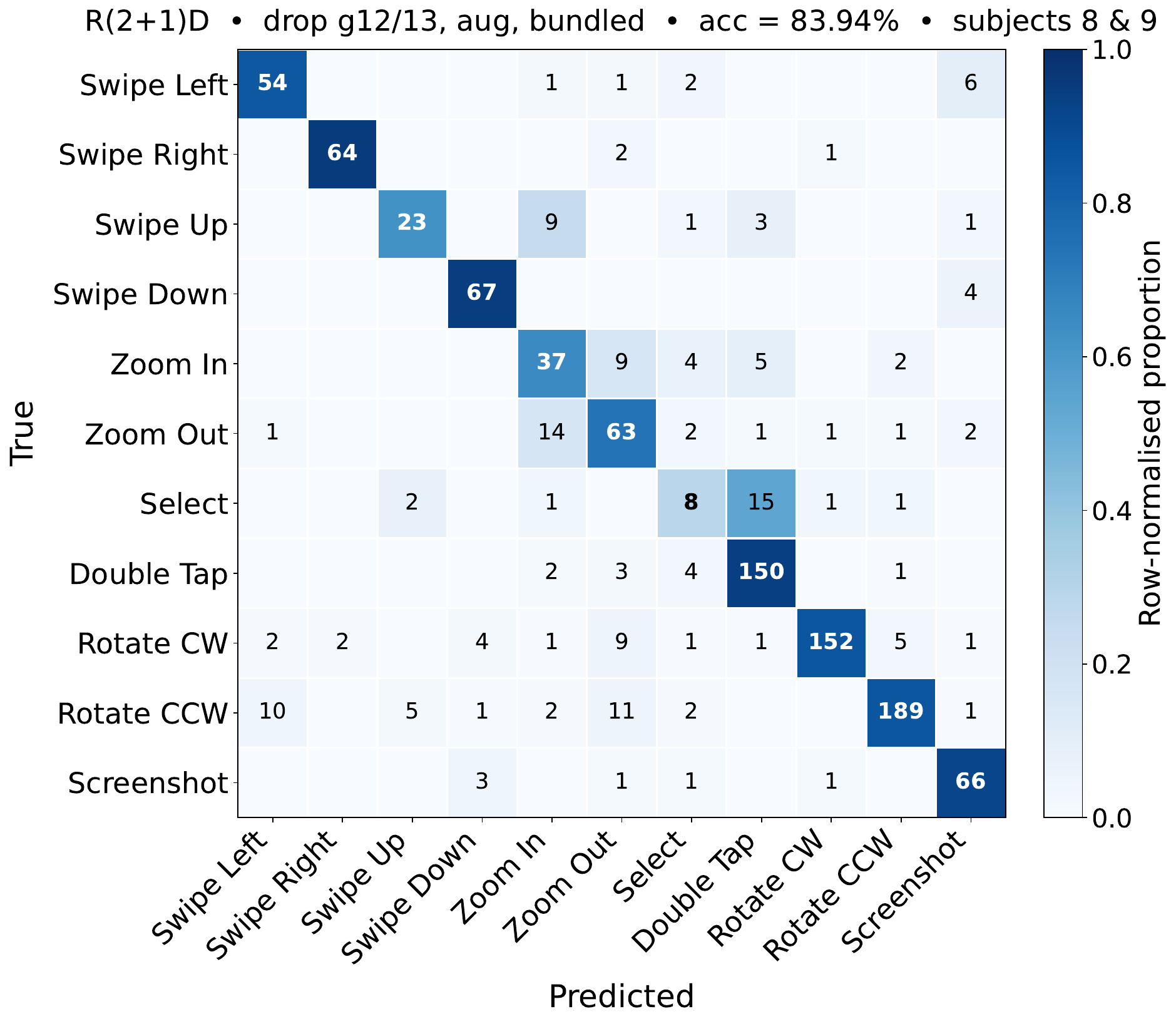}
    \caption{Row-normalised confusion matrix for the best model (R(2+1)D, drop g12/13, aug, bundled; test acc.\ 83.94\%, subjects 8 and 9). Diagonal entries show correctly classified counts; off-diagonal entries reveal the primary confusion pairs: Zoom In/Out mutual confusion, Select~$\to$~Double Tap, and Swipe Up~$\to$~Zoom In.}
    \label{fig:confusion_matrix}
\end{figure}

Figure~\ref{fig:confusion_matrix} shows the full confusion matrix for the best model. The dominant off-diagonal entries confirm the per-class analysis: Zoom In and Zoom Out exchange 14 and 9 samples respectively (the largest mutual confusion pair), Swipe Up is mistaken for Zoom In in 9 cases (both involve upward hand motion), 8 Select samples are misclassified as Double Tap (a spatially similar, more frequent gesture), and Rotate CCW leaks 10 samples into Swipe Left. All other off-diagonal entries are small ($\leq$6 samples), confirming that the remaining gesture pairs are well separated.

Table~\ref{tab:per_class} reports precision, recall, and F1 for each of the 11 retained gesture classes under the best model. The results cluster into three groups. \textbf{High-performing gestures} (F1\,$>{0.87}$): Swipe Right, Swipe Down, Rotate CW, Rotate CCW, Double Tap, and Screenshot. These share a strong, directionally asymmetric event flow that produces distinctive spatio-temporal signatures, and most also benefit from high support counts (160--221 samples for the rotation classes). \textbf{Mid-tier gestures} (F1\,0.70--0.87): Swipe Left and Swipe Up. Swipe Left is slightly weaker than Swipe Right, likely due to hand laterality effects in the egocentric view. Swipe Up suffers primarily from low support (37 samples), as indicated by its high precision (0.821) relative to its recall (0.622): the model is conservative in predicting this class due to insufficient training examples. \textbf{Challenging gestures} (F1\,$<{0.70}$): Zoom In, Zoom Out, and Select. The zoom pair produce spatially similar, radially symmetric event patterns that differ only in the direction of finger motion; the model confuses them with each other and with related gestures. Select, the weakest class (F1\,=\,0.333), involves a subtle finger-tap motion with minimal spatial extent, generating sparse events that are difficult to distinguish from background activity, and its low support (28 samples) further limits the model's ability to learn a robust representation.

\begin{table}[!t]
\centering
\caption{Per-class results for the best model (R(2+1)D, drop g12/13, aug, bundled; test accuracy 83.94\%).}
\label{tab:per_class}
\begin{tabular}{lcccc}
\toprule
\textbf{Gesture} & \textbf{Precision} & \textbf{Recall} & \textbf{F1} & \textbf{Support} \\
\midrule
Swipe Left     & 0.806 & 0.844 & 0.824 & 64  \\
Swipe Right    & 0.970 & 0.955 & 0.962 & 67  \\
Swipe Up       & 0.821 & 0.622 & 0.708 & 37  \\
Swipe Down     & 0.870 & 0.944 & 0.905 & 71  \\
Zoom In        & 0.561 & 0.649 & 0.602 & 57  \\
Zoom Out       & 0.630 & 0.741 & 0.681 & 85  \\
Select         & 0.400 & 0.286 & 0.333 & 28  \\
Double Tap     & 0.833 & 0.938 & 0.882 & 160 \\
Rotate CW      & 0.987 & 0.854 & 0.916 & 178 \\
Rotate CCW     & 0.918 & 0.855 & 0.885 & 221 \\
Screenshot     & 0.868 & 0.917 & 0.892 & 72  \\
\midrule
\textbf{Weighted avg} & & & \textbf{0.840} & 1040 \\
\textbf{Macro avg}    & & & \textbf{0.781} & 1040 \\
\textbf{mAP Score}    & & & \textbf{...} & 1040 \\
\bottomrule
\end{tabular}
\end{table}

Precision and recall imbalances offer additional diagnostic information. For Zoom In and Zoom Out, both precision and recall are moderate, indicating bidirectional confusion between the two classes. For Select, precision (0.400) substantially exceeds recall (0.286), suggesting the model rarely predicts this class but is reasonably accurate when it does—consistent with a high-confidence, low-sensitivity decision boundary shaped by severe class imbalance.

\subsection{Qualitative Examples}

\begin{figure}[!t]
\centering
\setlength{\tabcolsep}{1.5pt}

{\small\textit{Correct predictions}}\\[2pt]
\begin{tabular}{cccccc}
\includegraphics[width=0.155\linewidth]{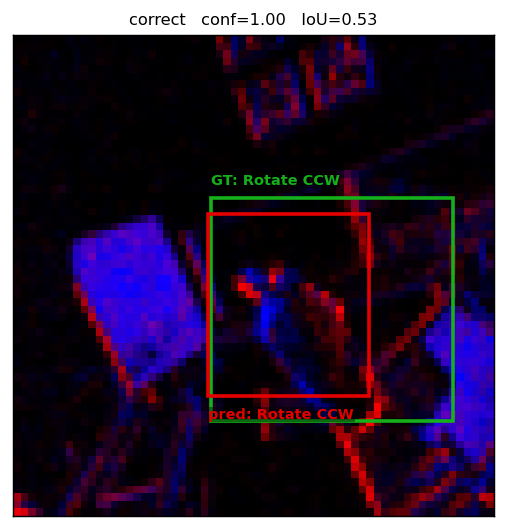} &
\includegraphics[width=0.155\linewidth]{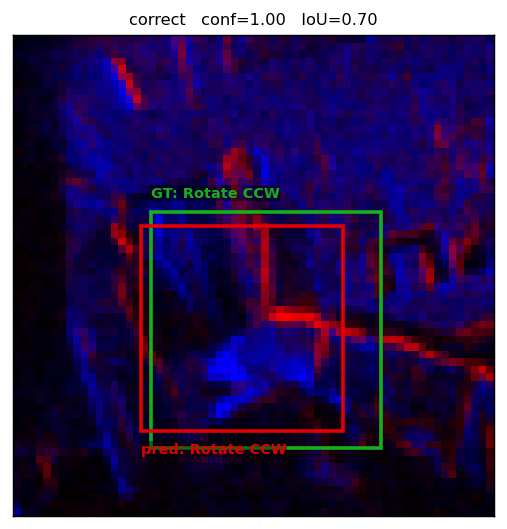} &
\includegraphics[width=0.155\linewidth]{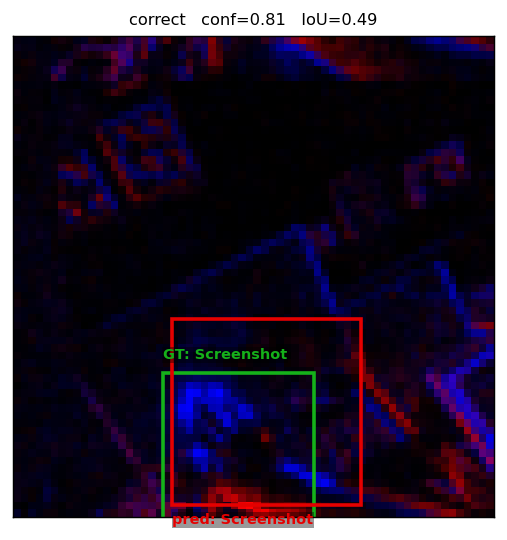} &
\includegraphics[width=0.155\linewidth]{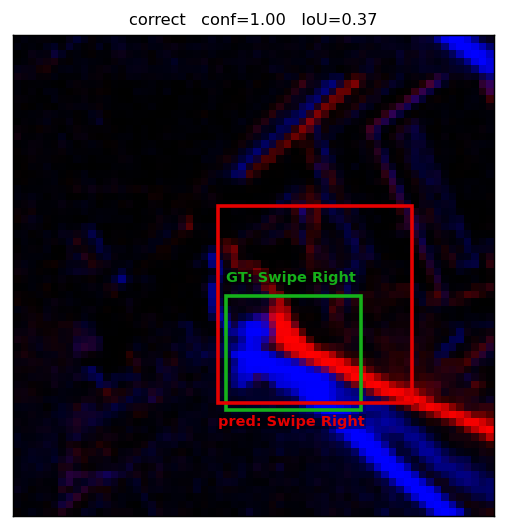} &
\includegraphics[width=0.155\linewidth]{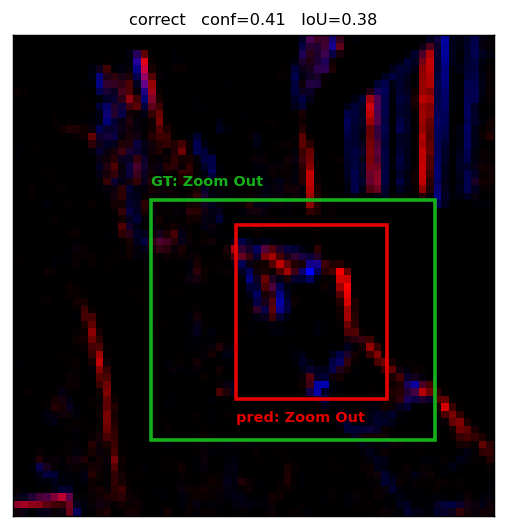} &
\includegraphics[width=0.155\linewidth]{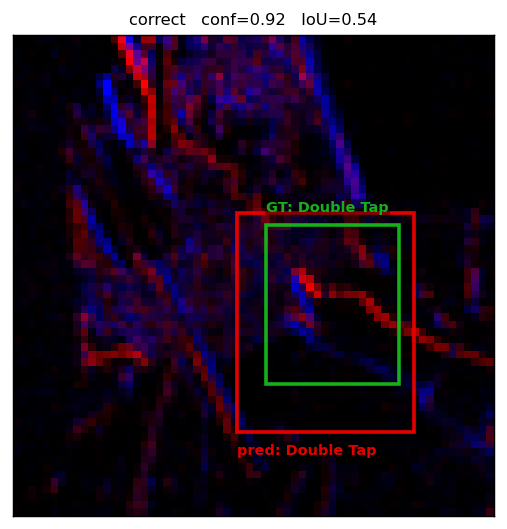}
\end{tabular}

\vspace{8pt}

{\small\textit{Misclassifications}}\\[2pt]
\begin{tabular}{cccccc}
\includegraphics[width=0.155\linewidth]{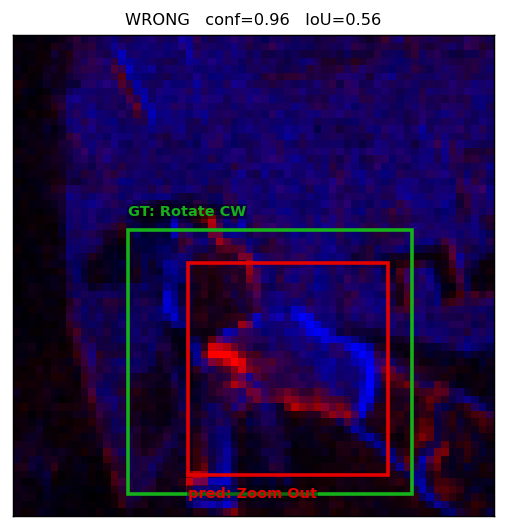} &
\includegraphics[width=0.155\linewidth]{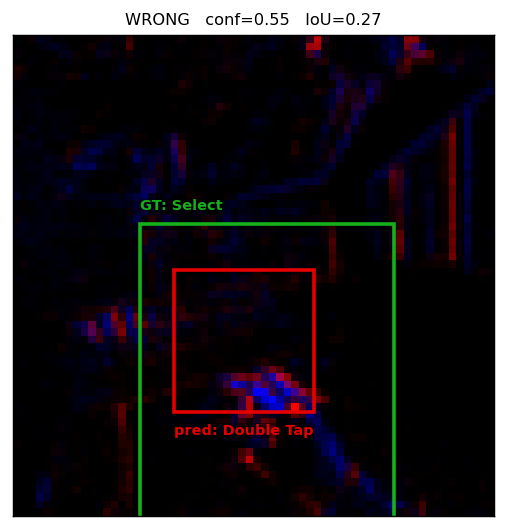} &
\includegraphics[width=0.155\linewidth]{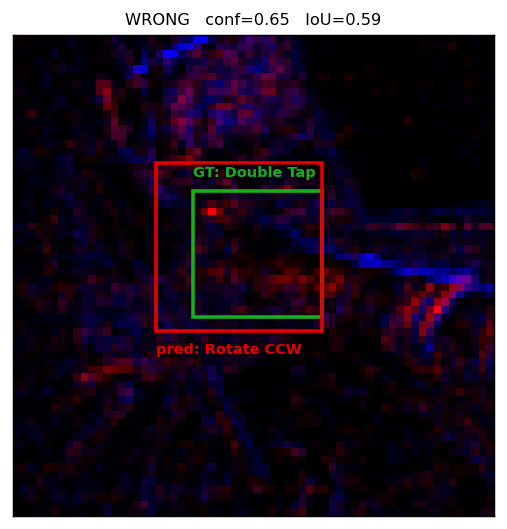} &
\includegraphics[width=0.155\linewidth]{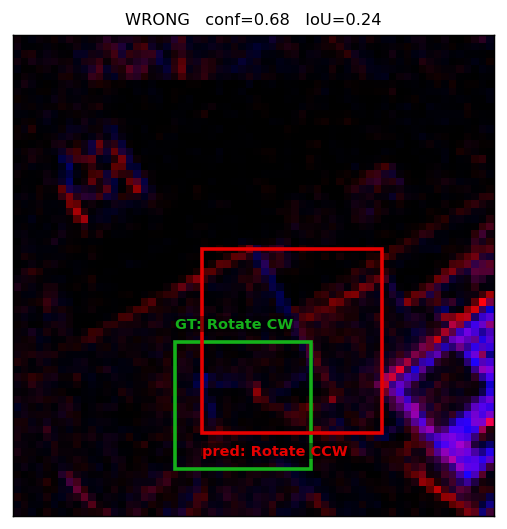} &
\includegraphics[width=0.155\linewidth]{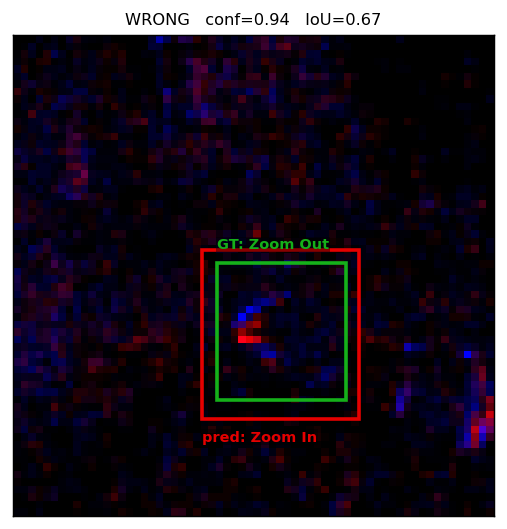} &
\includegraphics[width=0.155\linewidth]{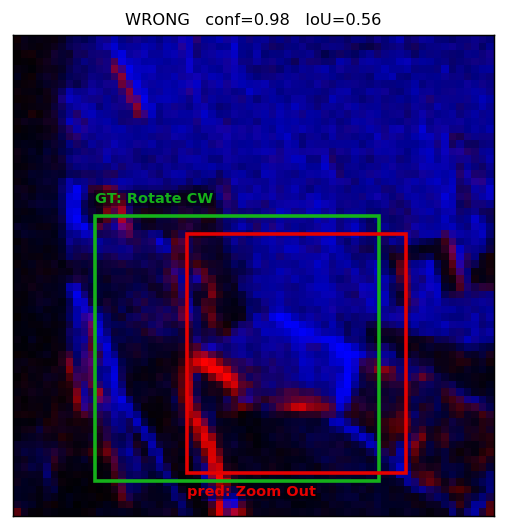}
\end{tabular}

\caption{Representative inference examples from the test set (subjects 8 and 9, best R(2+1)D model). 
\textit{Top row:} correctly classified samples. 
\textit{Bottom row:} misclassified samples, illustrating the main gesture confusion pairs discussed in the per-class analysis.}
\label{fig:inference_examples}
\end{figure}

Figure~\ref{fig:inference_examples} shows representative inference examples drawn from the test set. The top two rows present correctly classified samples across a range of gesture classes; the bottom two rows present misclassified samples.

%% file: 05_c_results_system.tex
\subsection{End-To-End System Evaluation}
\label{section:results_system}

\begin{figure}[!t]
    \centering
    \includegraphics[
        trim={0 0 0cm 0},
        clip,
        width=\linewidth
    ]{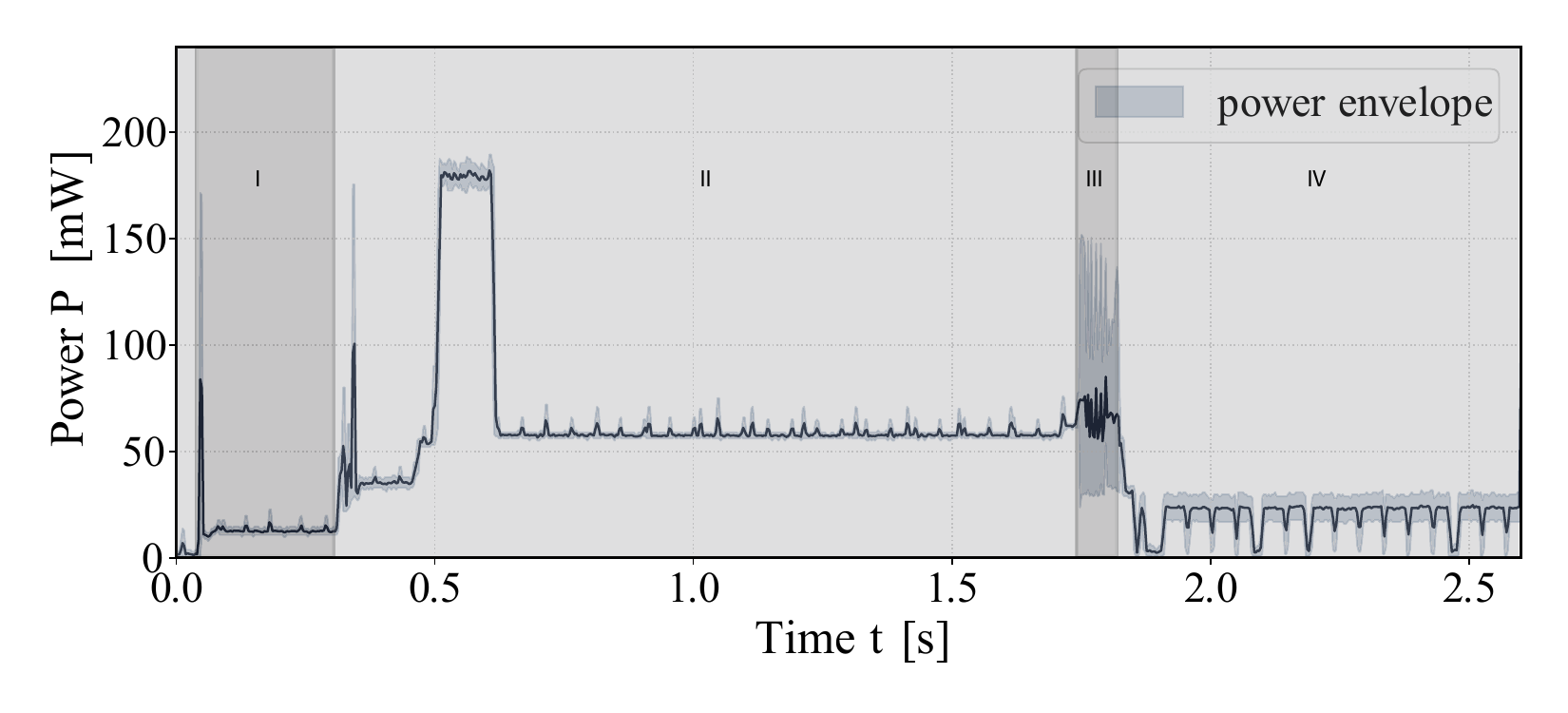}
    \caption{Measured power trace for a complete end-to-end inference cycle. Solid line, visualized the mean power per \SI{3.1}{ms} windows; power envelope shows instantaneous min-max range. Four phases are visible: (I) system cold start and GAP9 initialisation ($\approx$\,14.6\,mW average, with transient spikes during peripheral bring-up); (II) camera power-up, featuring a brief $\approx$\,180\,mW plateau as the GENX320 completes its cold-start sequence; (III) sustained vision pipeline execution at $\approx$\,67.4\,mW average, with periodic peaks corresponding to individual 78.3\,ms inference cycles; and (IV) BLE data transmission and deinitialization at $\approx$\,20.8\,mW average, showing duty-cycled power pulses as the nRF5340 advertises, connects, and transfers the result to a receiver application.}
    \label{fig:power_graph}
\end{figure}

Table~\ref{tab:gap_measurement_results} presents the per-stage execution time, average power, and energy consumption measured across a complete end-to-end inference cycle. \Cref{fig:power_graph} shows the corresponding power trace, which captures all four phases of operation: system initialization, sustained vision pipeline execution, and BLE transmission with deinitialization.

\begin{table}[h]
 \caption{Overview of cold start, execution time, and energy consumption across the stages of the image processing pipeline executed on the GAP9 \gls{MCU}.}
 \label{tab:gap_measurement_results}
 \centering
 \resizebox{\linewidth}{!}{%
 \begin{tabular}{lSSS}
  \toprule
   & $\bm{t}$\textbf{ [ms]} & $\bm{\overline{P}}$\textbf{ [mW]} & $\bm{E}$\textbf{ [mJ]} \\
  \midrule
   (I) Cold Start GAP9 & 287 & 14.6 & 4.1 \\
   (II) Cold Start GENX320 + Acquisition & 1420 & 65.6 & 92.9 \\
   (III) Inference & 78.3 & 67.4 & 5.2 \\
   (IV) BLE Transmission$^1$ & 778 & 20.8 & 16.0 \\
  \bottomrule
  \multicolumn{4}{l}{$^1$ Transmission of the event-frame and inference results via BLE.}
 \end{tabular}
 }
\end{table}

\textbf{Latency and throughput.}
 Model inference accounts for 78.3\,ms per frame (Phase~III)—the dominant bottleneck, slightly exceeding the 75.2\,ms inter-frame period at 13.3\,Hz and placing the GAP9 at near-full duty cycle, with image acquisition executing concurrently with the tail of the previous inference.

\textbf{Power trace analysis.}
The power trace in \cref{fig:power_graph} makes the four operational phases visually distinct. In \textbf{Phase~I} (cold start, $t \approx 2.7$--$3.0$\,s), the system draws approximately 14.6\,mW on average, with brief transient spikes reaching up to 170\,mW during peripheral initialisation. In \textbf{Phase~II} (camera bring-up, $t \approx 3.0$--$4.4$\,s), the GENX320 event camera requires a $\approx$\,1.4\,s cold-start sequence, visible as a sustained plateau near 180\,mW before the sensor begins generating event histograms. In \textbf{Phase~III} (continuous inference, $t \approx 4.4$--$5.5$\,s), the system enters the steady-state vision pipeline loop. The trace settles at $\approx$\,67.4\,mW average with periodic spikes corresponding to individual 78.3\,ms inference cycles peaking at $\approx$\,96\,mW during model execution on the GAP9. The regularity of these spikes at 13.3\,Hz confirms stable, uninterrupted pipeline throughput. In \textbf{Phase~IV} (BLE transmission and deinitialization, $t \approx 5.5$--$6.3$\,s), the system drops sharply to $\approx$\,20.8\,mW average after inference concludes, with duty-cycled power pulses as the nRF5340 coordinator advertises, establishes a BLE connection, and transmits the gesture prediction and bounding box to a paired receiver. This phase demonstrates the power benefit of the two-domain architecture: offloading wireless communication entirely to the nRF5340 allows the GAP9 and cameras to power down while data is transferred, reducing system power by more than $3\times$ compared to the active inference state.

\textbf{Energy breakdown and startup overhead.}
The one-time initialisation cost of 97.0\,mJ (Phases~I and~II: 4.1 + 92.9\,mJ) is 18.7$\times$ the energy of a single inference cycle (5.2\,mJ, Phase~III). For a session of $N$ consecutive inference cycles, the total energy is approximately $E_\text{total}(N) = 97.0 + N \times 5.2$\,mJ, giving an amortised per-inference cost that asymptotically approaches 5.2\,mJ for large $N$. At 13.3\,Hz, 110 inference cycles—corresponding to roughly 8.3\,s of active sensing—already reduce the amortised startup fraction below 15\%. This favours continuous, session-based operation over frequent cold-start cycling and motivates event-driven duty cycling at the session level: the nRF5340 remains always-on as a low-power watchdog, waking the performance domain only when sustained hand activity is detected, thereby avoiding repeated cold-start penalties.

\textbf{Runtime estimate.}
Model inference (Phase~III) consumes 5.2\,mJ per cycle at a steady-state average power of 67.4\,mW. At this operating point, a \qty{200}{\milli\ampere\hour} (\qty{774}{\milli\watt\hour}) battery sustains 11.5\,hours of continuous gesture recognition. 

%% file: 06_discussion.tex
\section{Discussion}
\label{section:discussion}

The modular \gls{FPC} interposer is the main architectural advantage of OpenGlass: it decouples sensor selection from the main board, allowing cameras to be swapped or extended without a full PCB redesign—unlike existing open-hardware platforms that fix the sensing modality. Combined with the \textsc{GAP9}'s NE16 accelerator and sub-byte quantisation support, the platform enables on-device inference workloads infeasible on typical \textsc{Cortex-M4} systems. It achieves 11.5\,h of continuous operation at 13.3\,Hz while running all \gls{ML} locally; event-driven duty cycling could extend this toward multi-day runtime.

R(2+1)D's superior performance stems from its factorised spatiotemporal design, which matches the structure of polarity-separated event histograms. The gap between the bundled hold-out result (83.94\,\%) and 6-fold cross-validation ($\approx$74--75\,\%) indicates that the held-out subjects are reasonably representative; the cross-validated figure is therefore the more conservative estimate. Symmetric and spatially compact gestures (\textit{Select}, \textit{Zoom In/Out}) remain the weakest classes due to subtle event signatures and limited dataset diversity.

Key limitations include the unused RGB camera (multimodal fusion is a natural next step), cross-subject generalisation bounded by the LynX dataset size, and the absence of prototype weight characterisation.

%% file: 07_conclusion.tex
\section{Conclusion}
\label{section:conclusion}

This paper presents OpenGlass, an open-source smart glasses platform for rapid prototyping of novel sensors and embedded \gls{ML} algorithms in an eyewear form factor. It addresses two key limitations of existing open hardware—fixed sensor modality and insufficient compute—through a modular \gls{FPC} interposer and a \textsc{GAP9} RISC-V \gls{SoC} capable of running vision inference within a sub-100\,mW envelope. A hardware--software co-designed power management strategy enables 11.5\,hours of continuous on-device \gls{ML} inference from a \qty{200}{\milli\ampere\hour} battery while processing all data locally.

As a reference application, an egocentric hand gesture recognition pipeline was evaluated on the LynX dataset. R(2+1)D achieves 83.94\% cross-subject accuracy (macro F1 = 0.781) with 78.3\,ms end-to-end inference latency on the GAP9 under a strict leave-two-subjects-out protocol. Temporal augmentation and removal of ambiguous classes contribute an 8.9 percentage-point accuracy gain. Six-fold cross-validation (74.6\% mean accuracy) shows that inter-subject variability remains the primary bottleneck. All hardware designs, firmware, and trained models are released as open source.

\section*{Acknowledgement \& Funding}

This work was supported by the Swiss National Science Foundation (\mbox{219943}) and Armasuisse. The authors would like to thank Dominik Mueller (hardware) and Varsha Jayaprakash (dataset).  Github Link: \url{https://github.com/ETH-PBL/OpenGlass}.
